\documentclass{article}
\usepackage{ijcai18}

\usepackage{times}  
\usepackage{helvet}  
\usepackage{courier}  
\usepackage{url}  
\usepackage{graphicx}  
\usepackage{amsopn}
\usepackage{amsmath}
\usepackage{amsfonts}
\usepackage{algorithm}
\usepackage[noend]{algpseudocode}
\usepackage{subcaption}
\usepackage{booktabs}
\usepackage{multirow}

\makeatletter
\def\BState{\State\hskip-\ALG@thistlm}
\makeatother

\usepackage{etoolbox}
\makeatletter
\patchcmd{\@verbatim}
  {\verbatim@font}
  {\verbatim@font\scriptsize}
  {}{}
\makeatother

\usepackage{amssymb}
\usepackage{pifont}
\usepackage{booktabs}
\usepackage[table,xcdraw]{xcolor}

\usepackage{array}
\newcolumntype{L}[1]{>{\raggedright\let\newline\\\arraybackslash\hspace{0pt}}m{#1}}
\newcolumntype{C}[1]{>{\centering\let\newline\\\arraybackslash\hspace{0pt}}m{#1}}
\newcolumntype{R}[1]{>{\raggedleft\let\newline\\\arraybackslash\hspace{0pt}}m{#1}}

\definecolor{green}{rgb}{0.1, 0.75, 0.0}

\begin{document}
%
\newcommand*\samethanks[1][\value{footnote}]{\footnotemark[#1]}

\title{{\em Explicability? Legibility? Predictability? Transparency? Privacy? Security?}\\ The Emerging Landscape of Interpretable Agent Behavior}

\author{
Tathagata Chakraborti$^1$ $\cdot$ Anagha Kulkarni$^1$ $\cdot$ Sarath Sreedharan$^1$\\
{\bf David E. Smith} $\cdot$ {\bf Subbarao Kambhampati}$^1$\\[0.5ex]
$^1$Arizona State University, Tempe 85281 USA \\[0.5ex]
\{tchakra2, akulka16, ssreedh3\}@asu.edu, david.smith@psresearch.xyz, rao@asu.edu
}

\maketitle
\begin{abstract}
There has been significant interest of late in generating behavior of agents that is interpretable to the human (observer) in the loop. 
However, the work in this area has typically lacked coherence on the topic,
with proposed solutions for ``explicable'', ``legible'', ``predictable'' and ``transparent'' planning with overlapping, and sometimes conflicting, semantics all aimed at some notion of understanding what intentions the observer will ascribe to an agent by
observing its behavior.
This is also true for the recent works on ``security'' and ``privacy'' of plans which are also trying to answer the same question, but from the opposite point of view -- i.e. when the agent is trying to hide instead of reveal its intentions.
This paper attempts to provide a workable taxonomy of relevant concepts in this exciting and emerging field of inquiry.
\end{abstract}

\begin{table*}[!th]
\tiny\hspace*{-20pt}
\centering
\begin{tabular}{@{}R{1.7cm}|rl|l|p{4cm}@{}}
\toprule
\rowcolor[HTML]{EFEFEF} 
\textbf{Concept} & \multicolumn{2}{l}{\cellcolor[HTML]{EFEFEF}\textbf{Setting / Agent Perspective}} & \textbf{Formulation / Existing Literature} & \textbf{Comments} \\ \midrule
& Agent & $\Pi^A = \langle \mathcal{M}^A, \mathcal{I}^A, \mathcal{G}^A \rangle, \chi^A, \Omega$ & Find: $\tilde{\pi}$ ($\pi$ in offline setting) & Find expected plan (satisfies observer model). \\ \cmidrule(lr){2-3}
\textbf{Explicability} & Observer & $\Pi^\Theta = \langle \mathcal{M}^\Theta, \mathcal{I}^\Theta, \mathcal{G}^\Theta \rangle, \chi^\Theta$ & Subject to: $_{\exists\pi\in\{\tilde{\pi}\}}\delta(\mathcal{I}^A, \pi, \chi^A) \models \mathcal{G}^A$ & \\ \cmidrule(lr){2-3}
& Target & Solve $\Pi^A$ with completion in $\Pi^\Theta$ & and $_{\exists\pi\in\{\tilde{\pi}\}, \langle o \rangle \models \tilde{\pi}}\delta(\mathcal{I}^\Theta, \pi, \chi^\Theta) \models \mathcal{G}^\Theta$ & 
Mainly concerns the plan prefix.
Note that the $\exists$ can be switched to $\forall$ to model a more pessimistic observer model that requires all possible completions be explicable.\\ \cmidrule(lr){4-5}
& \multicolumn{2}{r|}{\multirow{2}{*}{Related Work:}} & \cite{exp-yu} $\Pi^\Theta, \chi^\Theta$ unknown; $\mathcal{G}^A = \mathcal{G}^\Theta$; $\Omega: a \times s \mapsto a$ & $\Pi^\Theta, \chi^\Theta$ learned from human feedback in terms of a labeling scheme. \\ \cmidrule(lr){4-5}
& \multicolumn{2}{l|}{} & \cite{explicable-anagha} $\Pi^\Theta, \chi^\Theta$ unknown; $\mathcal{G}^A = \mathcal{G}^\Theta$; $\Omega: a \times s \mapsto a$ & $\Pi^\Theta, \chi^\Theta$ learned from human feedback in terms of plan distance $\Delta(\pi_1, \pi_2)$. \\ \cmidrule(lr){4-5}
& \multicolumn{2}{l|}{} & \cite{balancing} $\Pi^A \not= \Pi^\Theta$, $\chi^\Theta = O$, $\Omega: a \times s \mapsto a$ & Has the ability (via explanations) to deal with cases where 
$\not\exists\pi:\ \delta(\mathcal{I}^\Theta, \pi, \chi^\Theta) \models \mathcal{G}^\Theta$.\\ \midrule
& Agent & $\Pi^A = \langle \mathcal{M}^A, \mathcal{I}^A, \mathcal{G}^A \rangle, \chi^A, \Omega$ & Explicability + & Find most disambiguated (easy to predict) plan. \\ \cmidrule(lr){2-3}
\textbf{Predictability} & Observer & $\Pi^\Theta = \langle \mathcal{M}^\Theta, \mathcal{I}^\Theta, \mathcal{G}^\Theta \rangle, \chi^\Theta$ & $\min || \{ \pi\ |\ \pi\in\{\tilde{\pi}\}, \langle o \rangle \models \tilde{\pi}, \delta(\mathcal{I}^\Theta, \pi, \chi^\Theta) \models \mathcal{G}^\Theta \} ||$ & 
Mainly concerns the plan suffix. \\ \cmidrule(lr){2-5}
& Target & Solves $\Pi^A$ with fewest completions in $\Pi^\Theta$ & \cite{dragan2013legibility} $\Pi^\Theta$ implicit, $\chi^\Theta = O$, $\Omega: a \times s \mapsto a$ & 
Motion planning in continuous space.\\ \cmidrule(lr){4-5}
& \multicolumn{2}{r|}{\multirow{2}{*}{Related Work:}} & \cite{fisac2018generating} $\Pi^\Theta$ implicit, $\chi^\Theta = SF$, $\Omega: a \times s \mapsto o$ & 
Motion/semi-task planning in discrete space.\\ \cmidrule(lr){4-5}
& \multicolumn{2}{l|}{} & \cite{unified-anagha} $\Pi^A = \Pi^\Theta$, $\chi^\Theta = C$, $\Omega: a \times s \mapsto o$ & 
This work looks for \emph{m-similar} solutions in the offline case (for ``plan legibility'' or predictability) with similarity $d$ such that $|| \mathbb{S} || \geq m$ and $_{\forall\pi_1,\pi_2\in\mathbb{S}} \Delta(\pi_1,\pi_2) \leq d$, where $\mathbb{S} = \{ \pi\ |\ \delta(\mathcal{I}^\Theta, \pi, \chi^\Theta) \models \mathcal{G}^\Theta \}$ \\ \midrule
& Agent & $\Pi^A = \langle \mathcal{M}^A, \mathcal{I}^A, \mathcal{G}^A \rangle, \chi^A, \Omega$ & Find: $\tilde{\pi}$ ($\pi$ in offline setting) & Find plans that disambiguate possible goals.\\ \cmidrule(lr){2-3}
& Observer & $\Pi^\Theta = \langle \mathcal{M}^\Theta, \mathcal{I}^\Theta, \{\mathcal{G}^\Theta\} \rangle \equiv \{\Pi^\Theta_i\}, \chi^\Theta$  & Subject to: $_{\exists\pi\in\{\tilde{\pi}\}}\delta(\mathcal{I}^A, \pi, \chi^A) \models \mathcal{G}^A$ & \\ \cmidrule(lr){2-3}
\textbf{Legibility or Transparency} & Target & Solve $\Pi^A$ and least number of $\Pi^\Theta_i$s & and $\min ||\{ g\ |\ g \in \{\mathcal{G}^\Theta\} \wedge _{\exists\pi\in\{\tilde{\pi}\}, \langle o \rangle \models \tilde{\pi}}\delta(\mathcal{I}^\Theta, \pi, \chi^\Theta) \models g \}||$ &
Property of the goal. $\mathcal{G}^A$ may not be in $\{\mathcal{G}^\Theta\}$ as long as there is a mapping between them. \\ \cmidrule(lr){4-5}
& \multicolumn{2}{r|}{\multirow{2}{*}{Related Work:}} & \cite{dragan2013legibility} $\Pi^\Theta$ implicit, $\chi^\Theta = O$, $\Omega: a \times s \mapsto a$ & \\ \cmidrule(lr){4-5}
& \multicolumn{2}{l|}{} & \cite{macnally2018action} $\Pi^A=\Pi^\Theta$, $\chi^\Theta = O$, $\Omega: a \times s \mapsto a$ & \\ \cmidrule(lr){4-5}
& \multicolumn{2}{l|}{} & \cite{unified-anagha} $\Pi^A = \Pi^\Theta$, $\chi^\Theta = C$, $\Omega: a \times s \mapsto o$ & 
This work specifically looks for \emph{j-legible} solutions in the offline sense such that $|| \{ g\ |\ \delta(\mathcal{I}^\Theta, \pi, \chi^\Theta) \models g \} || \leq j$ \\ \bottomrule
\end{tabular}
\caption{Summary of Concepts (cooperative setting).}
\label{dave-1}
\vspace{-5pt}
\end{table*}

\subsection*{Introduction}

There has been significant interest in the robotics and planning community of late in developing algorithms that can generate behavior of agents that is interpretable to the human (observer) in the loop.
This notion of interpretability can be in terms of goals, plans or even rewards that the observer is able to ascribe to the agent based on observations of the latter.
While interpretability remains a significant challenge\footnote{As emphasized in the \emph{Roadmap for U.S. Robotics} \cite{christensen2009roadmap} 
-- {\em ``humans must be able to read and recognize agent activities in order to interpret the agent's understanding''}. 
} in developing human-aware AI agents, such as assistive agents, the work in this area has typically lacked coherence on the topic from the community as a whole, even if not in the research agenda of different research groups \cite{chakraborti2017ai,Dragan17,macnally2018action}, per se. 
Indeed, a quick scan of the existing literature reveals algorithms for ``explicable'', ``legible'', ``predictable'' and ``transparent'' planning with overlapping, and sometimes conflicting, semantics.
The same can be said of a parallel thread of work on the ``deception'', ``privacy'' and
``security'' of plans.
This paper thus attempts 
to provide a workable taxonomy of relevant concepts that can hopefully provide some clarity and guidance to future researchers looking to work on the topic.

The rest of the paper is organized as follows: 
We will first introduce a general framework for 
describing problems in the space of ``plan interpretability'' and outline how existing works
have addressed different aspects of this problem in
cooperative settings.
We will then turn the tables and explore their
complementary manifestations in adversarial settings.
Finally, we will end with a discussion on 
gaps in the proposed framework that are yet to be
explored in existing literature.

\subsection*{Model differences with the Observer}

The key challenge in generating interpretable behavior is the ability to account for the model of the observer.
This can be summarized as follows --

\begin{itemize}
\item An agent's actions may be uninterpretable 
when it does not conform to the expectations or predictions 
engendered by the observer model.
Thus, the agent, to plan for interpretable behavior, must not only consider its own model but also the observer model and the differences thereof.
\cite{chakraborti2017ai,Dragan17}
\end{itemize}


\begin{figure*}
    \centering
    \begin{subfigure}[b]{0.32\textwidth}
	    \centering
        \includegraphics[width=\textwidth]{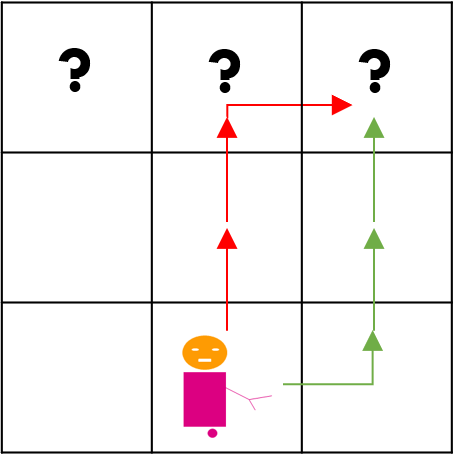}
        \caption{Plan legibility / transparency.}
        \label{fig:1}
    \end{subfigure}
    \hfill 
    \begin{subfigure}[b]{0.32\textwidth}
	    \centering
        \includegraphics[width=\textwidth]{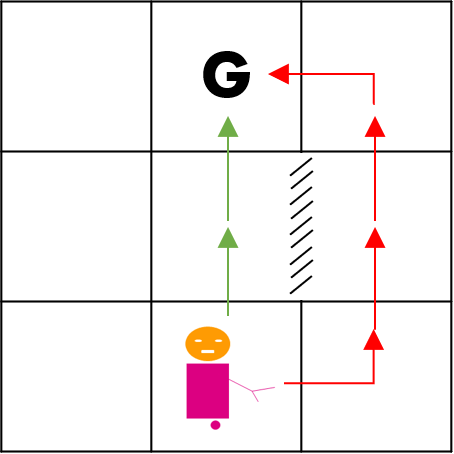}
        \caption{Plan explicability.}
        \label{fig:2}
    \end{subfigure}
    \hfill 
    \begin{subfigure}[b]{0.32\textwidth}
	    \centering
        \includegraphics[width=\textwidth]{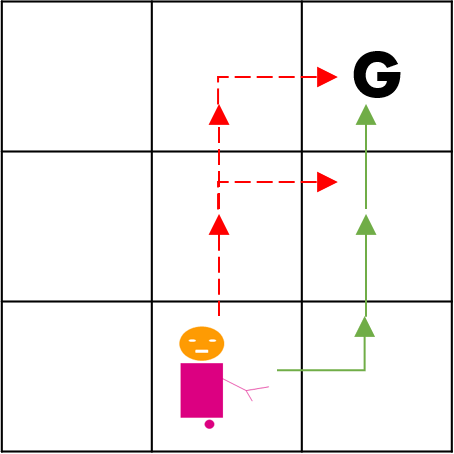}
        \caption{Plan predictability.}
        \label{fig:3}
    \end{subfigure}
\caption{
A simple illustration of the differences between plan explicability, legibility and predictability.
In this Gridworld, the agent can travel across cells, but cannot go backwards.
Figure \ref{fig:1} illustrates a legible plan (\textcolor{green}{green}) in the presence of 3 possible goals of the agent, marked with {\bf ?}s.
The \textcolor{red}{red} plan is not legible since all three goals are likely in its initial stages.
In the parlance of transparent planning, the first action in the \textcolor{green}{green} plan can constitute a transparent plan (having conveyed the goal).
Figure \ref{fig:2} illustrates an explicable plan (\textcolor{green}{green}) which goes straight to the goal {\bf G} as we would expect. 
The \textcolor{red}{red} plan may be more favorable to the agent due to its internal constraints (the arm sticking out might hit the wall), but is inexplicable (i.e. sub-optimal) in the observer's model. 
Finally, Figure \ref{fig:3} illustrates a predictable plan (\textcolor{green}{green})
since there is only one possible plan after it performs the first action.
In the parlance of {\em t-predictability}, this is a 1-predictable plan.
The \textcolor{red}{red} plans fail to disambiguate among two possible completions of the plan.
Note that all the plans shown in Figure \ref{fig:3} are explicable (optimal in the observer's model) but only one of them is predictable -- i.e. explicable plans may not be predictable.
Similarly, in Figure \ref{fig:2}, the \textcolor{red}{red} plan is predictable after the first action (even though not optimal, since there is only one likely completion) but not explicable -- i.e. predictable plans may not be explicable.
Without a prefix in Figure \ref{fig:2}, the \textcolor{green}{green} plan is the only predictable plan.
}
 	\label{fig:fig}
\end{figure*}

This ``model'' can include the beliefs or state information of the agent, its goals and intentions, its capabilities or even its reward function.
It can also include the observation model as well as the computational capability of the observer.
A misunderstanding or mismatch on any of those accounts will mean that the plan or policy, as expected by the observer (given their cognitive capabilities), will not be the same as that computed by the agent,
and will thus be difficult to interpret from the observer's point of view.
We will outline in the rest of this writeup how existing work on the topic
addresses one or more of these contributing factors, 
especially the goals and plans\footnote{It is useful to note
at this point that in this particular paper, we talk of behavior and
plan in the same breath. 
In general, behavior can be seen as a particular
instantiation of a plan or policy (which, in its general form, can have loops, contingencies, abstractions, etc.) 
or a policy. However, most of the works
surveyed here have used the term plan to primarily refer to behavior.
We will also stick to that convention -- i.e. all the discussion here
is confined to behaviors observed or ascribed to the agent by the
observer in the loop.}
ascribed to the agent by an observer.


Table \ref{dave-1} formalizes these considerations in the modeling of the agent $\mathcal{A}$ and the observer $\Theta$ in terms of --

\begin{itemize}
\item {\bf Planning Problem:} $\Pi = \langle \textnormal{Domain Theory} = \mathcal{M}, \textnormal{Current State} = \mathcal{I}, \textnormal{Goal State} = \mathcal{G}\rangle$ 
\item
{\bf Plan:} $\pi$ is a solution to the planning problem $\Pi$ emitting an observation sequence $\langle o \rangle \models\pi$. 
$\tilde{\pi}$ is a partial plan whose completion set is denoted by $\{\tilde{\pi}\}$. 
\item
{\bf Computational Model:} $\chi \in \{ S = \textnormal{Sound}, SF = \textnormal{Satisficing}, O = \textnormal{Optimal}, C = \textnormal{Complete}\}$
\item
{\bf Completion Model:} $\delta(s, \pi, \chi) \mapsto \hat{s}$ captures whether a state $s'$ is reachable from the state $s$ following a plan $\pi$ subject to the computation model (e.g. $\delta(\mathcal{I}, \pi, O) \mapsto \mathcal{G}$ implies $\pi$ is an optimal solution to $\Pi$); 
\item
{\bf Observation Model:} $\Omega: a \times s \mapsto o$
associates a token emitted for a particular action and next state pair. 
\end{itemize}

\subsection*{Interpretability? Plans versus Goals}

An agent model (and the corresponding observer model) 
thus accounts for their beliefs, goals, capabilities and even computation power. 
In such a formulation, the notion of {\em completion} is intrinsically related to the interpretability question --
the completion of a plan in that model is {\em equivalent} to whether they are interpretable given the assumptions on the model and computation power of the the observer.
The exact nature of the interpretation task may vary.
Most of the distinctions surrounding the interpretability\footnote{Explicability, legibility and predictability of plans is a spectrum, i.e. one plan can have more ``$X$''-ability than another. In the rest of the discussion, unless otherwise mentioned, we will refer to the end of that spectrum, whenever such a plan exists, (e.g. most explicable plan) when we mention an explicable, legible or predictable plan.}
of agent behavior deals with the disambiguation of {\em goals} versus {\em plans} \cite{dragan2013legibility} from the point of view of the observer.

\vspace{-15pt}
\paragraph{\em Explicability} 

We begin with ``plan explicability'' as introduced in \cite{balancing,zhang2016plan,exp-yu,explicable-anagha}.

\begin{quote}\em
Explicability measures how close a plan is to the expectations of the observer, given a known goal. 
\end{quote}

Thus the objective of explicability is to be in
the set of solutions to the observer's understanding of a planning problem.
In Table \ref{dave-1}, the explicable plan is  
one that has a completion in both the agent and the observer model.
The first constraint requires that the solution solves the agent's
planning problem while the latter requires that there exists a plan satisfying the emitted observations that enables a completion in the observer model -- e.g. the plan looks optimal to the observer as in \cite{balancing}.
When the observer model is not known \cite{exp-yu,explicable-anagha}, as is most often the case, the completion in the observer model is difficult to guarantee.
As such, {\em explicability is a spectrum}, where closer to completed plans in the observer model can be deemed to be more explicable.\\

\vspace{-10pt}
\paragraph{\em Predictability.} 

Plan predictability, on the other hand, looks for non-ambiguous completions of a plan prefix \cite{dragan2013legibility,fisac2018generating,unified-anagha}.

\begin{quote}\em
Plan predictability reduces ambiguity over possible plans, given a goal.
\end{quote}

Table \ref{dave-1} highlights this distinction with the additional minimization term over the cardinality of the possible plan set 
(that satisfies the emitted observations) 
with completions in the observer model.
This makes it clear that predictability is, again, a spectrum and -- 

\begin{quote}\em
An explicable plan can be unpredictable. 
\end{quote}

An example would be when there are multiple explicable plans, i.e. many completions in the observer model, so that there is still work to be done in making sure that the observer can anticipate which plan it is that the agent is going to execute. If this can be achieved, then that specific plan would be both explicable and predictable. 
Similarly --



\begin{quote}\em
A predictable plan (in the online setting) can be inexplicable in the offline setting. 
\end{quote}


This is possible when, given a prefix (during online plan execution), the observer can tell exactly what plan the agent is executing but the entire plan is still not one that s/he might expect it to (i.e. it does not follow the completion model of the observer).
For example, in \cite{fisac2018generating} the actions in the plan prefix of length $t$ can be arbitrary and inexplicable as long as the postfix is predictable. This is also true for transparent \cite{macnally2018action} plans as well. 
This phenomenon is readily seen in \cite{ppap} where the agent produces suboptimal plans that are easier to predict\footnote{Authors in \cite{fisac2018generating} use a fixed length of the plan prefix to generate predictable plan suffixes. 
In general, a planner can be allowed to determine this organically as done in \cite{ppap,macnally2018action}.
}. 
Figure \ref{fig:fig} provides another example.
More on this later in the discussion on online versus offline interactions.

\vspace{-10pt}
\paragraph{\em Legibility.} 
So far we have discussed explicability and predictability of plans under the condition of known goals only. 
Plan legibility, on the other hand, is defined as follows -- 

\begin{quote}\em
Plan legibility reduces ambiguity over possible goals that are being achieved.
\end{quote}

The observer model now includes a set of possible goals or equivalently 
a set of possible models parameterized by the goal, as shown in Table \ref{dave-1}. 
Now, in addition to solving the planning problem of the agent (first constraint as before), a legible solution requires that the set of observer models (or goals) where a plan satisfied by the emitted observations has completions (satisfies) is minimized.

The notion of legibility of goals has remained consistent across existing 
literature \cite{Dragan-2013-7732,dragan2013legibility,unified-anagha}.
and is equivalent to the notion of {\em transparency} of plans \cite{macnally2018action}.
To the best of our knowledge, plan explicability / predictability and legibility has 
not been considered together (i.e. with ambiguity over goals and plans simultaneously).

Interestingly, as Table \ref{dave-1} highlights, even though both predictability and explicability assume known goals, the goal known to the observer may not be the actual true goal
of the agent and yet plans may be predictable or explicable. 
For example, the agent could really be doing something else but also
achieve the expected goal with the desired behavior in the process.
The ability to communicate enables authors in \cite{balancing} to handle expectations under conditions of misunderstood goals as well. However, the notion of explicability remains identical as one of generating expected behavior with a shared understanding of the goal.

Similarly, for legibility to occur, there needs to be only 
some mapping between the agents goal and the possible goal set which
may not contain the real goal of the agent.

\subsection*{Online versus offline interactions.} 

The actual setup of the interaction 
-- i.e. online or offline -- 
makes a big difference to the 
explicability versus predictability discussion.
This is because explicability and predictability of a plan are {\em non-monotonic}, a plan prefix deemed inexplicable can become explicable with the execution of more actions
and vice versa, either due to the observer being an imperfect planner due to
computational limitations or due to implicit updates to the 
mental model based on the observations.
The online case of explicability can then be seen in terms of the 
plan prefix -- i.e. if its completion belongs to one of the explicable 
(completions in the observer model) or not.
On the other hand, the offline case does not exist for plan predictability,
which is a property of the plan suffix.
However, in the online case, before the execution starts (i.e. with no prefix)
a predictable plan has to be one of the explicable plans.
With a prefix, that may no longer be the case, as discussed above (this is considering
the definition explicability in the existing work on the entire plan).

Note that, similar to the concept of predictability, legibility of plans is more useful in the online setting since it may be easy to deduce the real goal from the final state after completion of the plan. 
Though, even in such cases, when the goals (which are not usually fully specified) are not mutually exclusive, legible plans can help. 
Like explicability and predictability, legibility also shares
the non-monotonicity property.


\subsection*{Motion versus Task Planning}

One of the biggest points of difference in many of these works is in the nature of the target domain – i.e. {\bf motion planning} \cite{Dragan-2013-7732,dragan2013legibility,dragan2015effects} versus {\bf task planning} \cite{exp-yu,explicable-anagha,zakershahrakinteractive,macnally2018action}. From the algorithmic perspective, this is simply differentiated in usual terms -- e.g. continuous versus discrete state variables. However, the notion of plan interpretability engenders additional challenges. This is because a reasonable mental model for motion planning\footnote{While this is true for path planning in general, complex trajectory plans of manipulators with high degrees of freedom might still require modeling of observer expectations.} can be assumed to be one that prefers shorter plans and thus need not be modeled explicitly (and thus does not need to be acquired or learned). 
For task planning in general, this is less straightforward. In fact, work on explicable task planning \cite{exp-yu,explicable-anagha,zakershahrakinteractive} has aimed to learn this implicit model using feedback from humans on the agent's behavior. 
A particularly instance of this is when these model are assumed to be identical \cite{macnally2018action,unified-anagha} (this is the case in motion planning, by default).

Given how humans can have vastly different expectations in the case of task planning, it is unclear how useful mental models learned from crowd feedback (as done in \cite{exp-yu,explicable-anagha,zakershahrakinteractive}) can be in the case of individual interactions.


\subsection{Computational Capability}

The discussion is, of course, contingent on the computational capability of the observer, as modeled in the 
completion function in Table \ref{dave-1}. 
There has been surprisingly little work to address this.
Authors in \cite{fisac2018generating} approximated the human model with Boltzmann noisy rationality.
Motion planning, again, can permit assumption of {\em ``top-$K$''} rationality while the 
computational model of the human is less clear in the task planning scenarios, i.e. domains with combinatorial properties 
(one can conceive of, for example, models of depth-bounded humans that constrains the space of plans in the mental model).
While almost all of the related work \cite{balancing,macnally2018action} has assumed
perfectly rational (super-)humans, models learned \cite{exp-yu,explicable-anagha} 
from feedback from human-subjects are likely to implicitly model computational limitations of the human mental model.

\subsection*{Discussion}


\paragraph{\em Learning the Observation Model.} 
The original work on explicability in task planning \cite{zhang2016plan,exp-yu} and subsequent works that build on it \cite{zakershahrakinteractive,gongbehavior} 
attempt to learn the observer model when it is unknown.
This is the only attempt to do so in the existing literature.
They postulate that the explicability\footnote{Authors in \cite{exp-yu} use ``explicability'' and ``predictability'' as measures towards achieving the same objective of producing plans closer to human expectation. This is somewhat confusing in the current context, though the notion of predictability used there for the disambiguation of the plan suffix remains consistent.} can be measured in terms of whether the human observer is able to associate higher level semantics to actions in the plan.
While this approach has its merits, it also arguably conflates explicability with predictability -- e.g. just because someone is able to assign task labels
to individual actions in a plan does not necessarily mean they would have expected that plan.


\vspace{-10pt}
\paragraph{\em Observability.} 
The concepts of explicability, predictability and legibility 
are intrinsically related to what is observable. 
In most of the existing work, the plan has been assumed to be completely observable. 
When this is not the case, the agent can try to ensure that unexpected actions are 
not observable and thus still be explicable. 
Interestingly most of the work in cooperative settings have worked with full observability while highlighting model differences. Later we will see that in the adversarial setting existing work mostly focuses on the observation model while assuming the rest of the agent's model is aligned with that of the observer.

\vspace{-10pt}
\paragraph{\em Longitudinal effects.} 
All of the work on the topic of interpretable behavior has, unfortunately,
revolved around single, and one-off, interactions and little attention has been
given to impact of evolving expectations in longer term interactions.
There is some reason to suspect that the need for explicable behavior will diminish
as humans become accustomed to the ``quirks'' of the agent. 
After all, to paraphrase George Bernard Shaw, {\em ``the world conforms to the unreasonable man''!}
This is, however, not a concern for legible and predictable behavior since, even
with complete model alignment, the topic of coordination remains relevant.

\vspace{-10pt}
\paragraph{\em Explanatory actions.} 
In recent work \cite{sarath555}, authors have explored the notion of 
{\em ``explanatory actions''} as actions that can have epistemic effects.
These are actions that can affect the observer model. 
Plans that are made explicable with the use of explanatory actions are, of course, never predictable -- 
i.e. one cannot {\em predict} that an explanatory action will occur during a behavior, but its presence can make the whole behavior explicable.
Thus, in this view, the set of possible explicable plans, not all
of them may be predictable.
But, as we discussed before, all the predictable plans at the start 
of plan execution have to be explicable.

\vspace{-10pt}
\paragraph{\em Human-agent Collaboration.} 
Note that most of the discussion till now has featured a human as a passive observer. 
However, in most scenarios, the human is likely to be a collaborator or, at the least,
their behavior is going to be contingent on that of the agent.
While explicability helps this cause, predictable behavior can arbitrarily (and negatively) effect the human when considered in isolation. 
Indeed, human factors studies of plan predictability versus legibility \cite{dragan2015effects} are consistent with this concern, demonstrating that legibility is more desirable in a collaborative setting.
Recent work \cite{zakershahrakinteractive} has started to take these considerations into account.

\vspace{-10pt}
\paragraph{\em On preference versus expectations.} 
There is considerable prior art on incorporating human preferences in robotic behavior, 
or plans in general. 
Indeed, the distinction between {\em preferences} and {\em expectations} is rather subtle.
The former can be seen as constraints imposed on the plan generation process if the agent
wants to contribute to the human's utility -- {\em ``What would Jesus want me to do?''} --
while the latter looks at how the agent can adapt its behavior in a manner that the human would
expect it to (as required by the human mental model) -- {\em ``What would Jesus expect me to do?''}.
As we mentioned before, in the case of motion planning, there is often no such distinction.
Even in the case of task planning -- for example, in ``human-aware'' planning where an agent decides not to vacuum while the elderly are asleep \cite{kockemann2014grandpa} --  sometimes it may be hard to identify where exactly the constraints lie, with preferences (``I don't want vacuuming while I am asleep'') or expectations (``I don't expect the agent to be designed to vacuum at odd hours'').
Ultimately this distinction might not make a difference algorithmically. 
The agent would need some process of performing 
multi-model argumentation 
(with its own model and the observer model)
during its planning process \cite{balancing}.

The lines do become even more blurred in experiments, unless carefully constructed, 
where human subjects are asked to label data with their expectation 
(i.e. how to ensure that they are not providing their preference instead?).
Unfortunately, the experimental design in \cite{exp-yu,explicable-anagha} is noticeably susceptible to this.

\begin{table*}[!th]
\tiny\hspace*{-20pt}
\centering
\begin{tabular}{@{}R{1.7cm}|rl|l|p{4cm}@{}}
\toprule
\rowcolor[HTML]{EFEFEF} 
\textbf{Concept} & \multicolumn{2}{l}{\cellcolor[HTML]{EFEFEF}\textbf{Setting / Agent Perspective}} & \textbf{Formulation / Existing Literature} & \textbf{Comments} \\ \midrule
& Agent & $\Pi^A = \langle \mathcal{M}^A, \mathcal{I}^A, \mathcal{G}^A \rangle, \chi^A, \Omega$ & Find: $\tilde{\pi}$ ($\pi$ in offline setting) & 
This is the inverse of the legibility problem.\\ \cmidrule(lr){2-3}
& Observer & $\Pi^\Theta = \langle \mathcal{M}^\Theta, \mathcal{I}^\Theta, \{\mathcal{G}^\Theta\} \rangle \equiv \{\Pi^\Theta_i\}, \chi^\Theta$ & Subject to: $_{\exists\pi\in\{\tilde{\pi}\}}\delta(\mathcal{I}^A, \pi, \chi^A) \models \mathcal{G}^A$ & \\ \cmidrule(lr){2-3}
& Target & Solve $\Pi^A$ and as many $\Pi^\Theta_i$ & and $\max ||\{ g\ |\ g \in \{\mathcal{G}^\Theta\} \wedge _{\exists\pi\in\{\tilde{\pi}\}, \langle o \rangle \models \tilde{\pi}}\delta(\mathcal{I}^\Theta, \pi, \chi^\Theta) \models g \}||$ &
A special case of {\bf simulation} \cite{masters2017deceptive} is $\delta(\mathcal{I}^\Theta, \pi, \chi^\Theta) \not\models \mathcal{G}^A$.
{\bf Deception} may or may not involve simulation. \\ \cmidrule(lr){4-5}
& \multicolumn{2}{r|}{\multirow{2}{*}{Related Work:}} & \cite{keren2015goal} $\Pi^A = \Pi^\Theta$, $\chi^\Theta = O$, $\Omega: a \times s \mapsto s$ & 
This is a special case of {\em k-ambiguity} \cite{unified-anagha} where $k=2$. Also, the solution is not secure \cite{secure-anagha} -- i.e. the real goal may not become the decoy if the algorithm is rerun with the decoy goal.\\ \cmidrule(lr){4-5}
\textbf{Goal-Obfuscation or Dissimulation or Privacy} & \multicolumn{2}{l|}{} & \cite{keren2016privacy} $\Pi^A = \Pi^\Theta$, $\chi^\Theta = O$, $\Omega: a \times s \mapsto o$ & 
Same as above (generalizes observation model). \\ \cmidrule(lr){4-5}
& \multicolumn{2}{l|}{} & \cite{masters2017deceptive} $\Pi^A = \Pi^\Theta$, $\chi^\Theta = O$, $\Omega: a \times s \mapsto a$ &
{\em Last Deceptive Point} (LDP) defined here (in the context of motion planning) has parallels to the notion of {\em equidistant states} in \cite{secure-anagha}. The latter deals with a general task planning setting. However, the latter deploys a heuristic which makes the planner incomplete. \\ \cmidrule(lr){4-5}
& \multicolumn{2}{l|}{} & \cite{unified-anagha} $\Pi^A = \Pi^\Theta$, $\chi^\Theta = C$, $\Omega: a \times s \mapsto o$ & 
This work specifically looks for \emph{k-ambiguous} solutions in the offline sense such that $|| \{ g\ |\ \delta(\mathcal{I}^\Theta, \pi, \chi^\Theta) \models g \} || \geq k$ \\ \midrule
& Agent & $\Pi^A = \langle \mathcal{M}^A, \mathcal{I}^A, \mathcal{G}^A \rangle, \chi^A, \Omega$ & Find: $\tilde{\pi}$ ($\pi$ in offline setting) & This is the inverse of the predictability problem.\\ \cmidrule(lr){2-3}
\textbf{Plan-Obfuscation} & Observer & $\Pi^\Theta = \langle \mathcal{M}^\Theta, \mathcal{I}^\Theta, \mathcal{G}^\Theta \rangle, \chi^\Theta$ & Subject to: $_{\exists\pi\in\{\tilde{\pi}\}}\delta(\mathcal{I}^A, \pi, \chi^A) \models \mathcal{G}^A$ & \\ \cmidrule(lr){2-3}
& Target & Solve $\Pi^A$ with most completions in $\Pi^G$ & and $\max || \{ \pi\ |\ \pi\in\{\tilde{\pi}\}, \langle o \rangle \models \tilde{\pi}, \delta(\mathcal{I}^\Theta, \pi, \chi^\Theta) \models \mathcal{G}^\Theta \} ||$ & \\ \cmidrule(lr){4-5}
& \multicolumn{2}{r|}{Related Work:} & \cite{unified-anagha} $\Pi^A = \Pi^\Theta$, $\chi^\Theta = C$, $\Omega: a \times s \mapsto o$ &
This work specifically looks for \emph{l-diverse} solutions in the offline sense such that $|| \{ \pi\ |\ \delta(\mathcal{I}^\Theta, \pi, \chi^\Theta) \models \mathcal{G}^\Theta \} || \geq l$\\ \midrule
& Agent & $\Pi^A = \langle \mathcal{M}^A, \mathcal{I}^A, \mathcal{G}^A \rangle, \chi^A, \Omega$ & Privacy + if $\langle o \rangle \models \hat{\pi}$ ($\pi$ in offline setting), then & \\ \cmidrule(lr){2-3}
\textbf{Security} & Observer & $\Pi^\Theta = \langle \mathcal{M}^\Theta, \mathcal{I}^\Theta, \{\mathcal{G}^\Theta\} \rangle \equiv \{\Pi^\Theta_i\}, \chi^\Theta$ & $\forall g \in \{ \mathcal{G}^\Theta \}$ : $_{\exists\pi\in\{\tilde{\pi}\}, \langle o \rangle \models \tilde{\pi}}\delta(\mathcal{I}^\Theta, \pi, \chi^\Theta) \models g$  & 
A privacy preserving planning algorithm is secure if it emits the same observation regardless of which goal it is run with.\\ \cmidrule(lr){2-5}
& Target & Find same solution for $\Pi^A$ and as many $\Pi^\Theta_i$ & \cite{secure-anagha} $\Pi^A = \Pi^\Theta$, $\chi^A = \neg{C}$, $\chi^\Theta = C$, $\Omega: a \times s \mapsto o$ & \cite{unified-anagha} can also allow this with a slight modification -- by generating an observation sequence that the agent wants to adhere to from the decoy goals. \\ \bottomrule
\end{tabular}
\caption{Summary of Concepts (adversarial setting).}
\label{dave-2}
\end{table*}

\begin{figure*}
    \centering
    \begin{subfigure}[b]{0.5\textwidth}
	    \centering
        \begin{subfigure}[b]{0.48\textwidth}
        \centering
        \includegraphics[width=\textwidth]{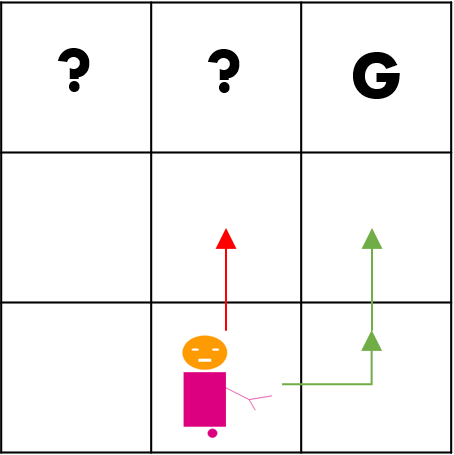}
        \caption{Simulation}
        \label{fig:2:1a}
        \end{subfigure}
    \hfill 
        \begin{subfigure}[b]{0.48\textwidth}
        \centering
        \includegraphics[width=\textwidth]{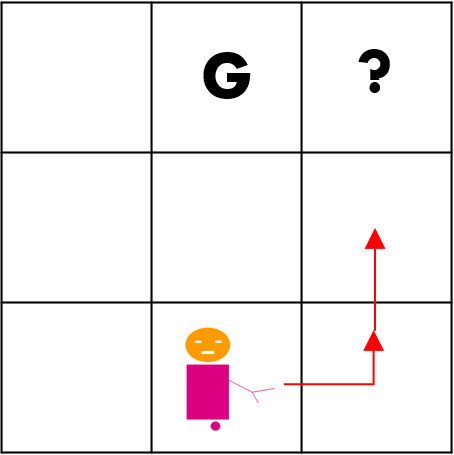}
        \caption{Dissimulation}
        \label{fig:2:1b}
        \end{subfigure}
        \caption{Deception: Simulation versus Dissimulation.}
        \label{fig:2:1}
    \end{subfigure}
    \qquad 
    \begin{subfigure}[b]{0.45\textwidth}
	    \centering
        \includegraphics[width=\textwidth]{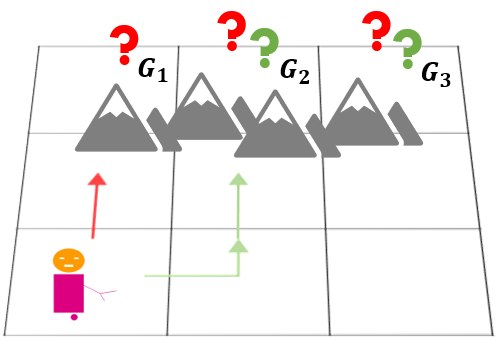}
        \caption{Obfuscation: Private versus Secure Plans.}
        \label{fig:2:2}
    \end{subfigure}
\caption{
A simple illustration of different goal obfuscation behaviors. 
Figure \ref{fig:2:1} shows different forms of deceptive behavior \protect\cite{masters2017deceptive} in \textcolor{red}{red} -- in simulation or ``hiding the truth'' (Figure \ref{fig:2:1a}) the agent could be going to either of the three possible goals while in dissimulation or ``showing the false'' (Figure \ref{fig:2:1b}) the likelihood of a decoy goal is strictly higher than that of the real goal.
The \textcolor{green}{green} plan in (Figure \ref{fig:2:1a}) is a truthful plan.
Figure \ref{fig:2:1} illustrates the difference between privacy and security of goal obfuscating plans. 
Here the observer cannot observe the actions of the agent in the first row of the grid due 
to occlusions. 
The \textcolor{red}{red} and \textcolor{green}{green} plans are both $2$-ambiguous \protect\cite{unified-anagha} and privacy preserving \protect\cite{keren2016privacy} -- 
the former allows for $\{G_1, G_2, G_3\}$ while the latter allows for $\{G_2, G_3\}$
as possible goal sets for the agent assumed to be rational.
However a secure algorithm cannot flip from the \textcolor{red}{red} to the 
\textcolor{green}{green} plan when rerun with $G_2$.
This is allowed under privacy preserving \protect\cite{keren2016privacy} and deceptive plans \protect\cite{masters2017deceptive} but not in secure plans \protect\cite{secure-anagha} -- 
i.e. the \textcolor{red}{red} plan is the only secure $2$-ambiguous solution.
}
\label{fig:fig:fig}
\end{figure*}

\section*{Turning the Tables}

So far we have talked about works that aim to reveal the 
intentions of the agent to an observer.
In a cruel twist of ``human awareness'', 
the agent can use the human mental model and/or the observation model to hide its intentions as well, in an adversarial sense.
In the following, we compare and contrast recent work in the planning community in this direction.
Many of the discussions, of course, carry over from our
treatise of plan explicability, predictability and legibility.

\vspace{-10pt}
\paragraph{\em Goals versus Plans}
Similar to the previous discussion on predictability/explicability versus legibility,
an agent can consider obfuscation of its goals and/or its plans.
The goal obfuscation problem is the inverse of the legibility problem,
while plan obfuscation is the inverse of the predictability problem discussed previously.
Also similar to the previous discussion, it is easy to see that
obfuscation of one (goal or plan) may not necessarily obfuscate the other.
Unsurprisingly, they can be viewed under a unified framework, as explored
recently in \cite{unified-anagha}.
Most of the existing work in this area has revolved around 
goal obfuscation (under the various names of privacy, deception and security)
as outlined in Table \ref{dave-2}.
Interestingly, these ideas has evolved out of two parallel threads of research
-- one \cite{keren2015goal,keren2016privacy,masters2017deceptive,macnally2018action} 
from the seminal work on {\em goal recognition design} \cite{keren2014goal} and the other \cite{unified-anagha,secure-anagha}
from the earlier work on plan explicability \cite{exp-yu,balancing,explicable-anagha}.
The connections between these diverse topics have hopefully
become apparent at this point.

\vspace{-10pt}
\paragraph{\em Motion versus Task Planning}

The distinction between motion and task planning again makes
an appearance in the techniques used to approach these problems
if not in the formulation of the concepts themselves.
Particularly, authors in \cite{masters2017deceptive} arrive at a 
computationally efficient proxy to the likelihood of possible
goals given a state and trajectory that is 
not necessarily available \cite{masters2017cost} 
to the task planning setting.
A similar computation used in \cite{secure-anagha}
turns out to be incomplete.

\vspace{-10pt}
\paragraph{\em Online versus Offline}

Intuitively, the obfuscation problem is much more appealing in the online case 
since some of the motivation in obfuscating plans (such as in evading a 
pursuit or escaping surveillance) is lost after the plan is done.
This is particularly the case for plan obfuscation in case of full observability, 
if not entirely true for goal obfuscation
(c.f. discussion on predictability and legibility in offline settings).
However, there is a rich set of problems to explore even in an offline setting 
\cite{unified-anagha,secure-anagha} 
once the observation model becomes noisy or incomplete.
An offline analysis also allows for the determination of 
prefixes, if not entire plans, that can preserve privacy \cite{keren2016privacy}. 
Most often, partial solutions (or fully obfuscating prefixes) is all
that is achievable.

\subsection*{The Observer and the Observation Model}

One of the most important variables in this thread of work
is the observation model.
The notion of obfuscation and privacy immediately follows from 
settings with partial observability \cite{keren2015goal,unified-anagha,secure-anagha}. 
However, as seen in \cite{masters2017deceptive,keren2014goal},
these concepts are still relevant to fully 
observable settings, especially in the online case.

The observer model also plays a critical role in this regard. 
Particularly, some of the surveyed works \cite{masters2017deceptive,keren2016privacy} 
have modeled a rational
(more powerful) observer where $P(\mathcal{G}\ |\ \langle o \rangle)$ follows the
distribution computed by a plan recognizer (i.e. $\chi^\Theta = O`$), while
others \cite{unified-anagha,secure-anagha} has considered
an observer with a complete planner, i.e. $P(\mathcal{G}\ |\ \langle o \rangle) = 1$
if a plan exists that satisfies the observation, $0$ otherwise (i.e. $\chi^\Theta = C$).
Interestingly, in contrast to explicability or legibility and predictability,
in the cases of goal or plan obfuscation,
none of the surveyed works have scratched the surface
of model differences between the agent and the observer beyond 
the particulars of the observation and computation model.

\subsection*{The Many Flavors of Security, Privacy, Obfuscation, (Dis-)Simulation and Deception}

The most interesting aspects of obfuscating behavior is 
the different notions of privacy, security and deception.
We highlight the salient differences in Figure \ref{fig:fig:fig}.

\vspace{-10pt}
\paragraph{\em Obfuscation and Privacy}

These are synonymous in the existing literature \cite{keren2016privacy,unified-anagha}
in attempting to make multiple goals likely or possible given a set of observations.
This is an inversion of the legibility problem --
as highlighted in Table \ref{dave-2}, the minimizing term has flipped.
Authors in \cite{keren2016privacy} look at pair-wise likelihood of goals in
maintaining privacy, while \cite{unified-anagha} provide a more general formulation
for ``k-ambiguous'' solutions.
This has implications on the security of the obfuscating algorithm, as we discuss below. 

\vspace{-10pt}
\paragraph{\em Security and Privacy}

The notion of security in plans \cite{secure-anagha} builds on the concept of 
``independence of inputs'' \cite{vstolba2017reveal,lindell2005secure}
which requires (in this specific context) that the planning algorithm return the same output or plan
regardless of which obfuscated goal it is run with as the real goal so
that an adversary cannot determine the real goal by rerunning 
the algorithm.
Thus -- 

\begin{quote}\em
A secure plan is always private. 
\end{quote}

This imposes an additional constraint to the privacy problem -- as highlighted in Table \ref{dave-2}, all possible goals must now lend to the same observation tokens.

\vspace{-10pt}
\paragraph{\em Obfuscation and Deception}

Finally, in goal (or plan) obfuscation, the primary goal
is to not reveal the true intentions, but not necessarily
actively mislead.
This distinction between simulation -- ``hiding the truth'' --
versus dissimulation -- ``showing the false'' -- was made
in \cite{masters2017deceptive}.
In the case of the latter, not only are multiple goals
likely given a plan prefix but a decoy goal is also
more likely than the real one.
Deception, in general, can include both.
It is clear from the discussion that -- 

\begin{quote}\em
A deceptive plan is always obfuscating, but may or may not be dissimulating.
\end{quote}

A more detailed discussion of this distinction can be found in \cite{masters2017deceptive}.




\section{Discussion}

In the following discussion, we make connections to
a parallel thread of work -- ``model reconciliation'' --
and outline possible directions for future work.

\subsection*{Communication and Model Reconciliation}

Most of the discussion in this paper has revolved around communication of 
intentions (goals or plans) implicitly using behavioral cues. 
In general, predictable or legible behavior can be seen as a special case of 
implicit signaling behavior \cite{gongbehavior} when communication is 
undesired. 
Foreshadowing certain actions (for example, through the medium of mixed 
reality \cite{ppap}) can considerably help the cause of predictability / legibility
and coordination in human-agent interaction.
The work on predictable \cite{fisac2018generating} or transparent \cite{macnally2018action} plans could have similarly deployed speech, stigmergic or, in general, communication actions in the plan prefix.
As mentioned before, recent work \cite{sarath555} provides
a unified formulation in terms of explanatory actions.


During communication, the agent must be able to address the root cause of inexplicability, i.e. it must be able to explicate parts of the model that differ from the human until they agree that its plan was, in fact, the best plan under the circumstances.
This process of explanation, referred to as a process of {\em model reconciliation},
has been of significant interest \cite{explain,explain-icaps,explain-ijcai,explain-hri} to the community recently.

{\em Particularly when the explicable plan is infeasible}, such communication remains the only option for the agent to achieve common ground with the human by, for example, expressing incapability \cite{raman2013sorry,raman2013towards,briggs2015sorry,kwon2018expressing}, communicating misunderstandings about its capabilities \cite{explain,balancing} or even lying \cite{lies} and augmenting new goals \cite{chenexplain}.
The latter works are certainly more relevant from the perspective of 
the second part of the paper which explores obfuscation of 
intentions instead of revealing them.
In fact, plan explicability and plan explanations form a delicate balancing act in ``human-aware planning'', 
as explored recently in \cite{balancing}. 
A concise survey of the model reconciliation process can be read in \cite{xaip}. 

\subsection{Further Generalizations}

In Tables \ref{dave-1} and \ref{dave-2} we provided a general framework for describing 
the different aspects of the plan interpretability problem.
The table also highlights gaps in the existing literature that
can lead to exciting avenues of research in future.
The model considered in Tables \ref{dave-1} and \ref{dave-2}, 
even though quite general in being able to classify 
the breadth of existing work on the topic,
does not quite capture the full scope of the 
plan interpretability. 
Below, we motivate a couple of generalizations to the 
framework presented in Tables \ref{dave-1} and \ref{dave-2}.
This was done intentionally so as not overly generalize
the overview which already captures all of the surveyed literature.

\subsubsection{Observation Model with Epistemic Effects}

The observation model used in Tables \ref{dave-1} and \ref{dave-2} is quite
general in being able to capture both partial as well as
noisy sensor models. This model has been used extensively
in the past \cite{geffner2013concise} as well as in 
many of the works covered in this survey; and provides
a particularly elegant sensor model while formulating
the planning problem for a single agent.
However, when considering an observer in the loop, one should be
cognizant of the effects of observations on the 
observer model -- i.e. epistemic effects of actions.
In recent work \cite{sarath555} this has been explored
in the context of implicit model updates on the part of
the observer by means of ``explanatory actions''.
One can conceive a more richer observation model
that captures such epistemic effects of the 
actions of an agent on the observer model.

\subsubsection{Preference Measure on Plan or Goal Set}

The notion of legibility and obfuscation \cite{unified-anagha,secure-anagha,masters2017deceptive,keren2016privacy}
has largely considered the computation of a {\em set} of plans or goals as the desired consequence of a behavior, 
with additional preferences on the cardinality of that set in certain cases (e.g. predictability). 
Interestingly, in the solution for plan-legibility or predictability, authors in \cite{unified-anagha}
look at ``l-diverse'' and ``m-similar'' solutions that can equally apply to the goal obfuscation and legibility
cases as well.
In general, the minimization or maximization term over the plan or goal sets in Tables \ref{dave-1} and \ref{dave-2}
can be replaced by a function over the preferences of the observer towards the agent's 
achievement (execution) of any particular goal (plan) in the possible goal (plan) set,
with cardinality being a special case of that function.
More on this below.

\subsubsection{An Active / Semi-Passive Observer}

All the work surveyed here consider a passive observer. 
The full scope of the interpretability problem is likely 
to include a more capable observer.
This can be a semi-passive observer -- i.e. one that 
can change the observation model only (in a sense reversal of the
``sensor cloaking'' problem explored in \cite{keren2016privacy}),
for example, to improved observability by going to higher ground
-- to a fully active observer with their own goals and actions,
with the ability to even assist or impede the agent from achieving its goals.
This is likely to effect the relative importance of agent behaviors
(e.g. is predictability more important than legibility in a collaborative setting? \cite{dragan2015effects})
and also effect the preference measure as discussed above (e.g. a surveillance scenario makes
certain behaviors in the completions set more important to recognize, and hence to obfuscate, than others).

\subsubsection{Unified Approach to Interpretable Behavior}

As we mentioned before, existing work
has only looked at the different notions of interpretable behavior in isolation. 
Designing these behaviors is likely to become more 
challenging as we consider the effects of one or more of these 
behaviors simultaneously. 
For example, what would it mean to be explicable or predictable
when there is ambiguity over the agent's goals?
A legible plan given a goal might be an explicable plan 
for another goal. 
From our previous discussion regarding the fact
that any of these behaviors can exist
with or without the other,
it will be interesting to see how they can exist simultaneously.
Further, given that some of these behaviors are predicated on
the notion of rationality on the agent model only (explicability) 
and others are not (legibility and predictability), 
it is unclear how the observer may be modeled once the belief of rationality
has been suspended (for example, due to inexplicable but legible behavior).

\subsubsection{Behaviors versus Plans}

Though we alluded to this distinction very briefly at the start of the paper, our discussion has mostly been confined to analysis of behaviors -- i.e. one particular observed instantiation of a plan or policy.
In particular, a plan -- which can be seen as a set of constraints on behavior -- engenders a {\em candidate set} of behaviors
\cite{kambhampati1996candidate} some of which may have certain interpretable properties while others may not.
However, this also means that an algorithm that can capture the 
``$X$''-ability of a plan can also do so for a particular behavior 
it models since in the worst case a behavior is also a plan that has 
a singular candidate completion.
A general treatment of a plan can be very useful in the offline setting -- e.g. in decision-support \cite{radar}
where human decision-makers are deliberating over possible plans with the support from an automated planner.
Unfortunately, interpretability of such plans has received 
very little attention beyond explanation generation \cite{smith2012planning,danmaga,borgo2018towards}. 

\section*{Conclusion}

In conclusion, we looked at a variety of 
interpretable behaviors of an agent 
which provides a rich set of directives to consider
while designing agents that can account for the 
observer model in their decision making processes.
We also saw how the ability to model and anticipate
interpretability of its own behavior can be dual-use --
i.e. the agent can use this to either reveal or obfuscate
its intentions to the observer.
We compared and contrasted existing literature that 
has tackled various aspects of this problem 
and provided a unified framework for precise specification 
of these (often confused) ideas.
We also highlighted gaps in existing work 
and directions for future research.
Finally, in this survey we have focused on the interpretability
of behavior only, and the role of privacy and obfuscation in
that context only.
There is a rich body of work in the planning community
that has explored these concepts in the context of {\em information
sharing} in multi-agent planning \cite{brafman2015privacy,vstolba2017reveal} 
that can provide additional insights towards a more general of formulation of 
privacy preservation and obfuscation in a joint planning scenario.

\bibliographystyle{named}
\bibliography{bib}

\end{document}